\documentclass[letterpaper, 10 pt, journal, twoside]{IEEEtran}

\IEEEoverridecommandlockouts                              

\usepackage{amsmath} 
\usepackage{amssymb}  
\usepackage{cite}
\usepackage{booktabs}
\usepackage[subnum]{cases}
\usepackage{mathtools}
\usepackage{balance}
\usepackage{color}
\usepackage{tabularx}
\usepackage{graphicx,dblfloatfix} 
\usepackage{listings}
\usepackage{textcomp}
\usepackage{url}
\usepackage{multirow}
\usepackage{algorithm}
\usepackage[noend]{algpseudocode}
\usepackage{ragged2e} 
\newcolumntype{Y}{>{\RaggedRight\arraybackslash}X} 

\usepackage{bm}
\usepackage{bbm}
\usepackage[colorinlistoftodos]{todonotes}
\usepackage{hyperref}
\usepackage[normalem]{ulem}
\usepackage{units}
\usepackage{threeparttable} 

\newcommand{\norm}[1]{\left\lVert#1\right\rVert}
\newcommand{\abs}[1]{|#1|}

\usepackage[printonlyused]{acronym}

\acrodef{RSL}{Robotic Systems Lab}
\acrodef{COM}{center of mass}
\acrodef{SQP}{sequential quadratic problem}
\acrodef{VMC}{virtual model controller}
\acrodef{w.r.t.}{with respect to}
\acrodef{DOF}{degrees of freedom}
\acrodef{ZMP}{zero moment point}
\acrodef{IMU}{inertial measurement unit}
\acrodef{COT}{cost of transport}
\acrodef{JPL}{Jet Propulsion Laboratory}
\acrodef{HAA}{hip adduction/abduction}
\acrodef{HFE}{hip flexion/extension}
\acrodef{KFE}{knee flexion/extension}
\acrodef{ZMP}{zero-moment point}
\acrodef{QP}{quadratic programming}
\acrodef{SQP}{sequential quadratic programming}
\acrodef{WBC}{whole-body controller}
\acrodef{HO}{hierarchical optimization}
\acrodef{NLP}{nonlinear programming}
\acrodef{MPC}{model predictive control}
\acrodef{TO}{trajectory optimization}
\acrodef{DARPA}{Defense Advanced Research Projects Agency}
\acrodef{JPL}{Jet Propulsion Laboratory}
\acrodef{RBDL}{Rigid Body Dynamics Library}

 \newcommand{\deleted}[1]{}

\title{
Rolling in the Deep -- Hybrid Locomotion for Wheeled-Legged Robots using Online Trajectory Optimization
}

\author{Marko Bjelonic$^{1}$, Prajish K. Sankar$^{2}$, C. Dario Bellicoso$^{3}$, Heike Vallery$^{4}$ and Marco Hutter$^{1}$
\thanks{Manuscript received: September, 10, 2019; Revised: December, 18, 2019; Accepted: January, 18, 2020.}
\thanks{This paper was recommended for publication by Editor Nikos Tsagarakis upon evaluation of the Associate Editor and Reviewers' comments. This work was supported in part by the Swiss National Science Foundation (SNF) through the National Centres of Competence in Research Robotics (NCCR Robotics) and Digital Fabrication (NCCR dfab). Besides, it has been conducted as part of ANYmal Research, a community to advance legged robotics.}
\thanks{Correspondence should be addressed to Marko Bjelonic.}
\thanks{$^{1}$ M. Bjelonic and M. Hutter are with the Robotic Systems Lab, ETH Z\"urich, 8092 Z\"urich, Switzerland.
{\tt\footnotesize marko.bjelonic@mavt.ethz.ch}}
\thanks{$^{2}$ P. K. Sankar is a student at the Faculty of Mechanical, Maritime and Materials Engineering, Delft University of Technology, 2628 CD Delft, Netherlands and was with the Robotic Systems Lab, ETH Z\"urich, 8092 Z\"urich, Switzerland at the time of this study.}
\thanks{$^{3}$ C. D. Bellicoso is with Boston Dynamics, 02451 Waltham, Massachusetts, United States and was with the Robotic Systems Lab, ETH Z\"urich, 8092 Z\"urich, Switzerland at the time of this study.}
\thanks{$^{4}$ H. Vallery is with the Faculty of Mechanical, Maritime and Materials Engineering, Delft University of Technology, 2628 CD Delft, Netherlands.}
\thanks{Digital Object Identifier (DOI): see top of this page.}
}

\begin{document}

\markboth{IEEE Robotics and Automation Letters. Preprint Version. Accepted January, 2020}{Bjelonic \MakeLowercase{\textit{et al.}}: Rolling in the Deep -- Hybrid Locomotion for Wheeled-Legged Robots using Online Trajectory Optimization}

\maketitle

\begin{abstract}
Wheeled-legged robots have the potential for highly agile and versatile locomotion. The combination of legs and wheels might be a solution for any real-world application requiring rapid, and long-distance mobility skills on challenging terrain. In this paper, we present an online trajectory optimization framework for wheeled quadrupedal robots capable of executing hybrid walking-driving locomotion strategies. By breaking down the optimization problem into a wheel and base trajectory planning, locomotion planning for high dimensional wheeled-legged robots becomes more tractable, can be solved in real-time on-board in a model predictive control fashion, and becomes robust against unpredicted disturbances. The reference motions are tracked by a hierarchical whole-body controller that sends torque commands to the robot. Our approach is verified on a quadrupedal robot with non-steerable wheels attached to its legs. The robot performs hybrid locomotion with a great variety of gait sequences on rough terrain. Besides, we validated the robotic platform at the Defense Advanced Research Projects Agency (DARPA) Subterranean Challenge, where the robot rapidly mapped, navigated and explored dynamic underground environments.
\end{abstract}


\section{INTRODUCTION}
\begin{figure}[t]
    \centering
    \includegraphics[width=\columnwidth]{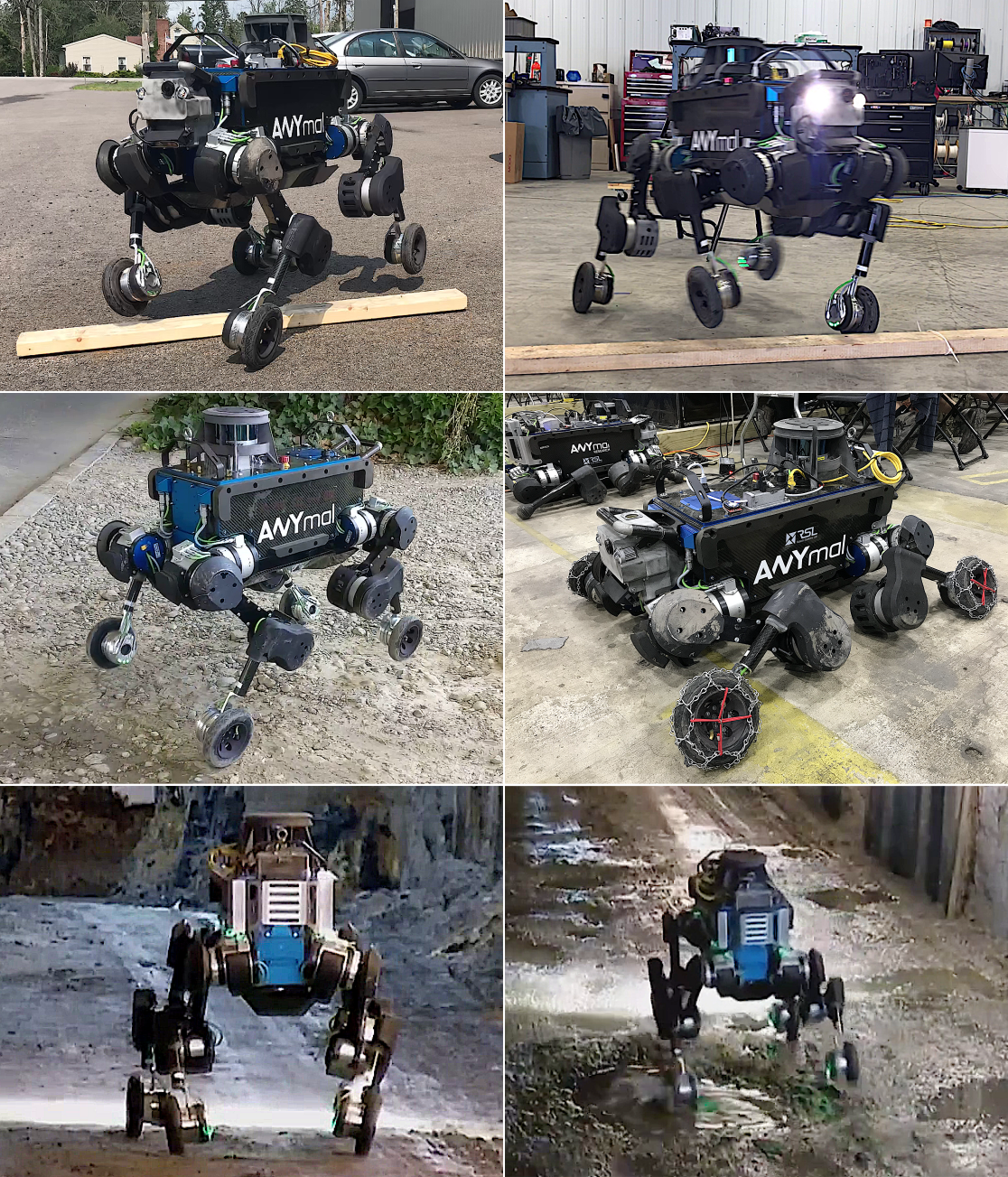}
     \caption{The fully torque-controlled quadrupedal robot ANYmal~\cite{hutter2017anymaljournal} equipped with four non-steerable, torque-controlled wheels. The robot is traversing over a wooden plank (top images), on rough terrain (left middle image). In addition, the robot rapidly maps, navigates and searches dynamic underground environments at the \ac{DARPA} Subterranean Challenge (lower images), and the robot's wheels are equipped with chains to traverse the muddy terrain (right middle image). A video can be found at \href{https://youtu.be/ukY0vyM-yfY}{https://youtu.be/ukY0vyM-yfY}.}
    \label{fig:anymal_on_wheels}
\end{figure}

\IEEEPARstart{L}{egged} 
robots offer the possibility of negotiating challenging environments and, thus, are versatile platforms for various types of terrains~\cite{bellicoso2018jfr}. In research and industry, there is an emphasis on replicating nature to improve the hardware design and algorithmic approach of robotic systems~\cite{eckert2015comparing,nyakatura2019reverse}. Even with extensive research, matching the locomotion skills of conventional legged robots to their natural counterparts remains elusive. In contrast, wheels offer a chance to extend some capabilities, particularly speed, of these legged robotic systems beyond those of their natural counterparts, which can be crucial for any task requiring rapid and long-distance mobility skills in challenging environments. With this motivation, the central contribution of this work involves locomotion planning on a wheeled-legged robot to perform dynamic hybrid\footnote{In our work, \emph{hybrid} locomotion denotes simultaneous walking and driving.} walking-driving motions on various terrains, as shown in Fig.~\ref{fig:anymal_on_wheels}.
\subsection{Related Work}
The online generation of optimal solutions for dynamic motions has been an active research area for conventional legged robots. Methods like \ac{TO} and \ac{MPC} are prevalent and recommended in the literature for aiding robots to be reactive against external disturbances and modeling errors. Finding control policies for performing walking motions in an articulated mobile robot is an involved task because of the system's many \ac{DOF} and its nonlinear dynamics. This demands substantial computational power and introduces the challenge of overcoming local minima, making on-the-fly computations hard.

In the literature concerning wheeled-legged robots, hybrid walking-driving motions are scarce. The focus is mostly on statically-stable driving motions where the legs are used for active suspension alone~\cite{reid2016actively,giordano2009kinematic,cordes2014active,giftthaler2017efficient,suzumura2014real,grand2010motion}. These applications do not show any instance of wheel lift-offs. Hence, sophisticated motion planning for the wheels is unnecessary and, therefore, usually skipped.

Agile motions over steps and stairs are demonstrated for the first time in our previous work~\cite{bjelonic2019keep}, where a hierarchical \ac{WBC} tracks the motion trajectories that include the rolling conditions associated with the wheels. The robot can execute walking and driving motions, but not simultaneously due to missing wheel trajectories over a receding horizon. As such, the robot needs to stop and switch to a pure walking mode to overcome obstacles. The work in \cite{viragh2019trajectory} extends the approach by computing base and wheel trajectories in a single optimization framework. This approach, however, decreases the update rate to \unit[50]{Hz}, and no hybrid walking-driving motions are shown on the real robot.

\emph{CENTAURO}, a wheeled-legged quadruped with a humanoid upper-body, performs a walking gait with automatic footstep placement using a linear \ac{MPC} framework~\cite{laurenzi2018quadrupedal}. The authors, however, only perform walking maneuvers without making use of the wheels. In contrast, the path planner in \cite{klamt2018planning} shows driving and walking motions in simulation without considering the robot's dynamics. Among the robots that employ hybrid walking-driving motions, Jet Propulsion Laboratory's (JPL) \emph{Robosimian} uses a \ac{TO} framework~\cite{bellegarda2019trajectory}, but for passive wheels and results are only shown in a simulation. \emph{Skaterbots}~\cite{geilinger2018skaterbots} provide a generalized approach to motion planning by solving a \ac{NLP} problem. This approach, however, is impractical to update online in a receding horizon fashion, i.e., in a \ac{MPC} fashion, due to excessive computational demand.

Given state of the art, we notice a research gap in trajectory generation methods for hybrid walking-driving motions on legged robots with actuated wheels, which can be both robust on various terrains and be used on-the-fly. Fortunately, research in traditional legged locomotion offers solutions to bridge this gap. The quadrupedal robot \emph{ANYmal} (without wheels) performs highly dynamic motions using \ac{MPC}~\cite{bellicoso2017dynamic,grandia2019feedback} and \ac{TO}~\cite{winkler2018gait,carius2019trajectory} approaches. Impressive results are shown by \emph{MIT Cheetah}, which performs blind locomotion over stairs~\cite{di2018dynamic} and jumps onto a desk with the height of \unit[0.76]{m}~\cite{nguyen2019optimized}. The quadrupedal robot \emph{HyQ} shows an online, dynamic foothold adaptation strategy based on visual feedback~\cite{magana2019fast}. Therefore, we conjecture that extending these approaches to wheeled-legged systems can aid in producing robust motions. 
\subsection{Contribution}
In our work, we present an online \ac{TO} framework for wheeled-legged robots capable of running in a \ac{MPC} fashion by breaking the problem down into separate wheel and base \ac{TO}s. The former takes the rolling constraints of the wheels into account, while the latter accounts for the robot's balance during locomotion using the idea of the \ac{ZMP}~\cite{vukobratovic2004zero}. A hierarchical \ac{WBC}~\cite{bjelonic2019keep} tracks these motions by computing torque commands for all joints. Our \emph{hybrid locomotion framework} extends the capabilities of wheeled-legged robots in the following ways:

1) Our framework is versatile over a wide variety of gaits, such as pure driving, statically stable gaits, dynamically stable gaits, and gaits with full-flight phases.

2) We generate wheel and base trajectories for hybrid walking-driving motions in the order of milliseconds. Thanks to these fast update rates, the resulting motions are robust against unpredicted disturbances, making real-world deployment of the robot feasible. Likewise, we demonstrate the performance of our system at the \ac{DARPA} Subterranean Challenge, where the robot autonomously maps, navigates and searches dynamic underground environments.
\section{MOTION PLANNING}
\label{sec:motionPlanning}
The whole-body motion planner is based on a task synergy approach~\cite{farshidian2017planning}, which decomposes the optimization problem into wheel and base \ac{TO}s. By breaking down the problem into these two tasks, we hypothesize that the issue of locomotion planning for high-dimensional (wheeled-)legged robots becomes more tractable. The optimization can be solved in real-time in a \ac{MPC} fashion, and with high update rates, the locomotion can cope with unforeseen disturbances.

\begin{figure}[t!]
    \centering
    \includegraphics[width=\columnwidth]{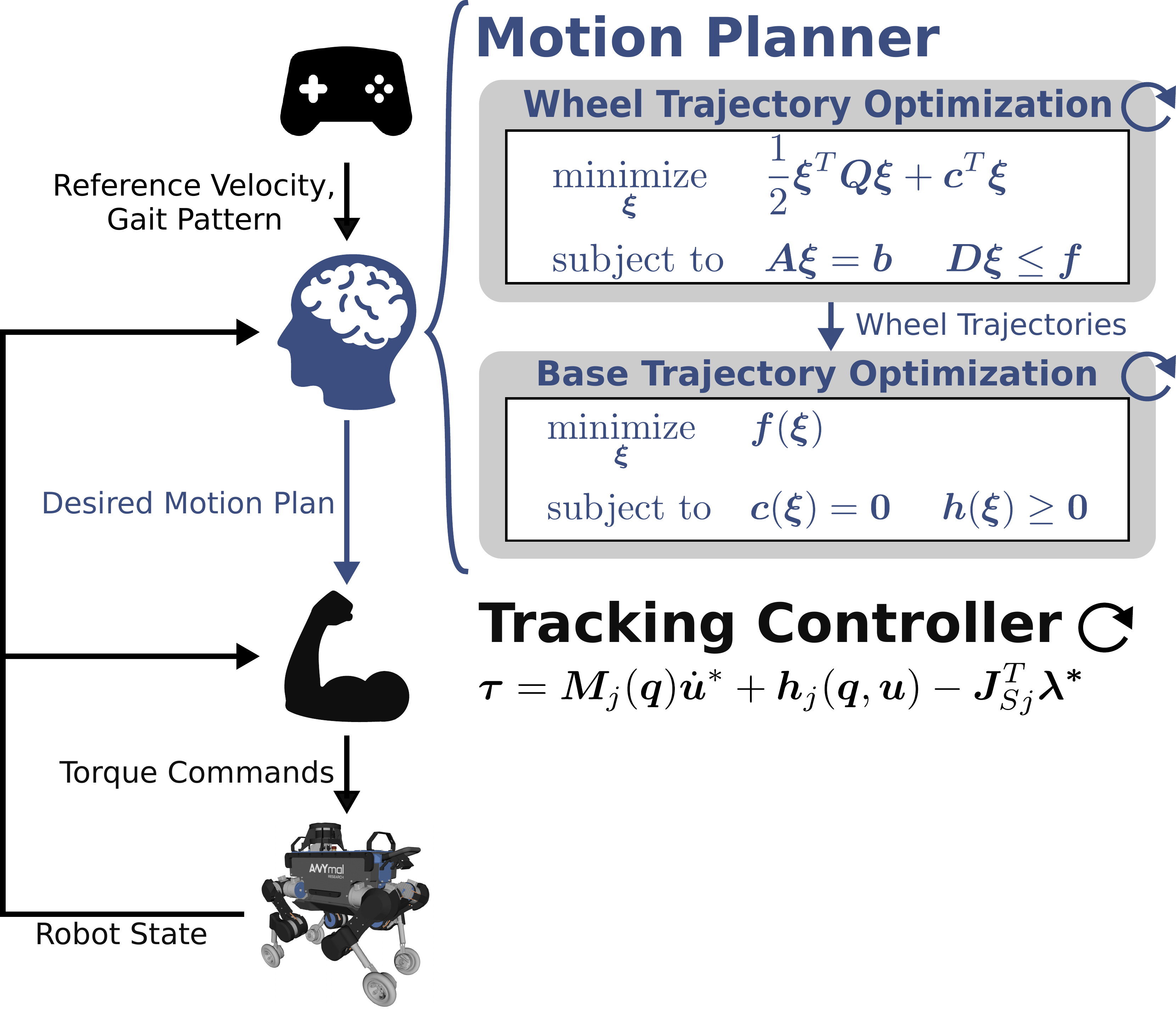}
    \caption{Overview of the motion planning and control structure. The motion planner is based on a \ac{ZMP} approach, which takes into account the optimized wheel trajectories and the state of the robot. The hierarchical \ac{WBC}, which optimizes the whole-body accelerations $\bm{\dot{u}^*}$ and contact forces $\bm{\dot{\lambda}^*}$, tracks the operational space references. Finally, torque references $\bm{\tau}$ are sent to the robot. The wheel \ac{TO}, base \ac{TO}, and \ac{WBC} can be parallelized due to the hierarchical structure.}
    \label{fig:motion_planner_overview}
\end{figure}

The main idea behind our approach is visualized in Fig.~\ref{fig:motion_planner_overview}. Given a fixed gait pattern and the reference velocities\footnote{The reference velocities are generated from an external source, e.g., an operator device, or a navigation planner.} \ac{w.r.t.} the robot's base frame $B$ as shown in Fig.~\ref{fig:wheel_trajectory}, i.e., the linear velocity vector of its \ac{COM} $\bm{v}_{\mathrm{ref}}$ and the angular velocity vector $\bm{\omega}_{\mathrm{ref}}=\begin{bmatrix}0 & 0 & \omega_{\mathrm{ref}}\end{bmatrix}^T$, desired motion plans are generated in two steps, where the wheel \ac{TO} is followed by a base \ac{TO} which satisfies the \ac{ZMP}~\cite{vukobratovic2004zero} stability criterion. The latter simplifies the system dynamics for motion planning of the \ac{COM} to enable real-time computations onboard. Finally, a controller tracks these motion plans by generating torque commands which are sent to the robot's motor drives. Due to this decomposition of the locomotion problem, the wheel \ac{TO}, the base \ac{TO}, and the tracking controller can run in parallel.

The following two sections discuss the main contribution of our work and show how the locomotion of the independent wheel and base \ac{TO}s are synchronized to generate feasible motion plans.
\section{WHEEL TRAJECTORY OPTIMIZATION}
\label{sec:wheelTrajectoryOptimization}
\begin{figure}[t!]
    \centering
    \includegraphics[width=0.9\columnwidth]{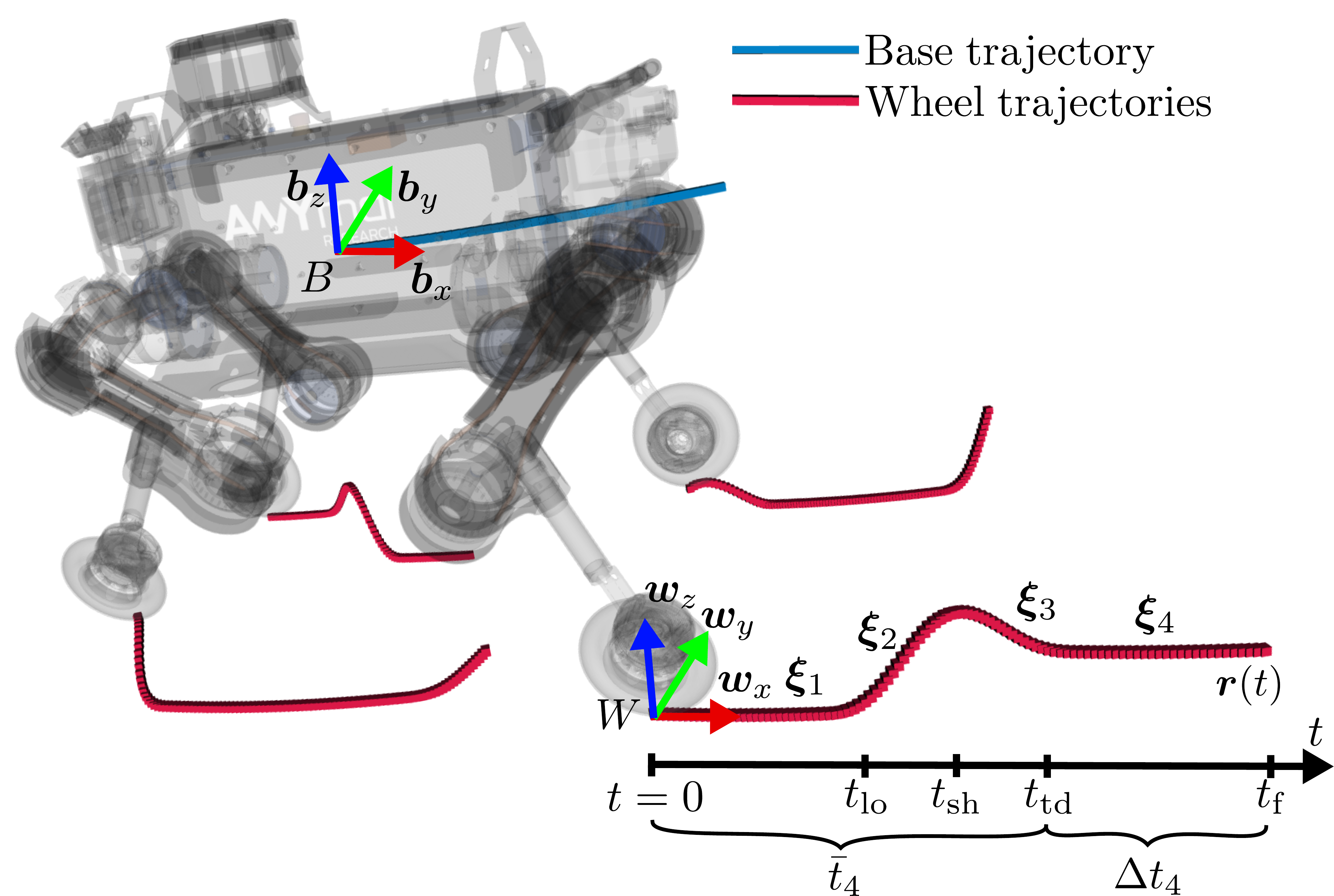}
    \caption{Timings and coordinate frames. The figure shows a sketch of the wheel and base trajectory. The wheel trajectories are optimized for each of the wheels separately and \ac{w.r.t.} the coordinate frame $W$ whose $z$-axis is aligned with the estimated terrain normal, and whose $x$-axis is perpendicular to the estimated terrain normal and aligned with the rolling direction of the wheel. The origin of $W$ is at the projection of the wheel's axis center on the terrain. We show exemplarily the wheel trajectory of the right front leg over a time horizon of one stride duration, which is composed of four splines. The lift-off time $t_{\mathrm{lo}}$, the time at maximum swing height $t_{\mathrm{sh}}$, the touch-down time $t_{\mathrm{td}}$, and the time horizon $t_\mathrm{f}$ are specified by a fixed gait pattern. The base trajectories are optimized \ac{w.r.t.} the coordinate frame $B$ whose origin is located at the robot's \ac{COM}, and whose orientation is equal to that of the frame $W$.}
    \label{fig:wheel_trajectory}
\end{figure}
We formulate the task of finding the wheel trajectories, i.e., the $x$, $y$ and $z$ trajectories \ac{w.r.t.} a wheel coordinate frame $W$ as illustrated in Fig.~\ref{fig:wheel_trajectory}, as a separate \ac{QP} problem for each of the wheels given by
\begin{equation}
\label{eq:quadraticProgram}
\begin{aligned}
& \underset{\bm{\xi}}{\text{minimize}}
& & \frac{1}{2} \bm{\xi}^T \bm{Q} \bm{\xi} + \bm{c}^T \bm{\xi}, \\
& \text{subject to}
& & \bm{A} \bm{\xi} = \bm{b}, \ \bm{D}\bm{\xi} \leq \bm{f},
\end{aligned}
\end{equation}
where $\bm{\xi}$ is the vector of optimization variables. The quadratic objective $\frac{1}{2} \bm{\xi}^T \bm{Q} \bm{\xi} + \bm{c}^T \bm{\xi}$ is minimized while respecting the linear equality $\bm{A} \bm{\xi} = \bm{b}$ and inequality  $\bm{D}\bm{\xi} \leq \bm{f}$ constraints. In the following, the parameterization of the optimization variable is presented, and we introduce each of the objectives, equality constraints and inequality constraints which form the optimization problem.
\subsection{Parameterization of Optimization Variables}
We describe the wheel trajectories as a sequence of connected splines. In our implementation, one spline is allocated for each of the two segments where the wheel is in contact with the ground, and two splines are used for describing the trajectory of the wheels in the air. Therefore, the total number of splines for one gait sequence is $n_{\mathrm{s}}=4$ (see Fig.~\ref{fig:wheel_trajectory}). These two types of trajectory segments, i.e., corresponding to leg in the air and contact, are defined by different parameterizations as described next.
\subsubsection{Wheel segments in air}
\label{sec:wheel_segments_in_air}
We parameterize each coordinate of the wheel trajectory in air as quintic splines. Thus, the position vector at spline segment $i$ is described by
\begin{equation}
\bm{r}(t) = 
\begin{bmatrix}
\bm{\eta}^T(t)& \bm{0}_{1\times6}& \bm{0}_{1\times6} \\
\bm{0}_{1\times6}& \bm{\eta}^T(t)&\bm{0}_{1\times6} \\
\bm{0}_{1\times6}&\bm{0}_{1\times6}&\bm{\eta}^T(t)
\end{bmatrix}
\begin{bmatrix}
\bm{\alpha}_{i,x} \\
\bm{\alpha}_{i,y} \\
\bm{\alpha}_{i,z}
\end{bmatrix} =
\bm{T}(t)\bm{\xi}_{i},
\end{equation}
where $\bm{\eta}^T(t)=\begin{bmatrix}t^5 & t^4 & t^3 & t^2 & t & 1\end{bmatrix}$ and $\bm{\alpha}_{i,*}\in\mathbb{R}^6$ contains the polynomial coefficients. Here, $t \in [\bar{t}_i,\bar{t}_i+\Delta t_i]$ describes the time interval of spline $i$ with a duration of $\Delta t_i$, where $\bar{t}_i$ is the sum of all the previous $(i-1)$ splines' durations (see the example of the fourth spline in Fig.~\ref{fig:wheel_trajectory}). We seek to optimize the polynomial coefficients for all coordinates of spline segment $i$ and hence contain them in the vector $ \bm{\xi}_{i} = \begin{bmatrix} \bm{\alpha}_{i,x}^T & \bm{\alpha}_{i,y}^T & \bm{\alpha}_{i,z}^T \end{bmatrix}^T \in \mathbb{R}^{18}$.
\subsubsection{Wheel segments in contact}
As shown in our previous work~\cite{viragh2019trajectory}, we employ a different parameterization for wheel segments in contact, such that they inherently capture the velocity constraints corresponding to the no-lateral-slip of the wheel. For this purpose, we represent the wheel's velocity in the $x$ coordinate of $W$, i.e., the rolling direction, as a quadratic polynomial. In contrast, the velocities of the remaining directions are set to zero. Thus, the velocity vector of the $i$-th spline is
\begin{equation} \label{eq:wheelSegmentVelocity}
\dot{\bm{r}}(t) = 
\begin{bmatrix}
1 & t & t^2 \\
0 & 0 & 0 \\
0 & 0 & 0
\end{bmatrix}
\begin{bmatrix}
\alpha_{i,0} \\
\alpha_{i,1} \\
\alpha_{i,2}
\end{bmatrix},
\end{equation}
and the position vector is obtained by integrating \ac{w.r.t.} $t$ and adding the initial position $x_{i}(\bar{t}_i)$ and $y_{i}(\bar{t}_i)$ of the trajectory as
\begin{equation}\label{eq:wheelSegmentSpline}
\bm{r}(t) = 
\begin{bmatrix}
x_{i}(\bar{t}_i) \\
y_{i}(\bar{t}_i) \\
0
\end{bmatrix} + 
\int\limits_{\bar{t}_i}^{\bar{t}_i+\Delta t_i} \bm{R}(t \bm{\omega}_{\mathrm{ref}}) \dot{\bm{r}}(t) \mathrm{d}t = \bm{T}(\omega_{\mathrm{ref}},t) \bm{\xi}_i,
\end{equation}
where the rotation matrix $\bm{R}(t \bm{\omega}_{\mathrm{ref}})$ describes the change in the wheel's orientation caused by the reference yaw rate, i.e., the vector $\bm{\omega}_{\mathrm{ref}}=\begin{bmatrix} 0 & 0 & \omega_{\mathrm{ref}} \end{bmatrix}^T$. By assuming a constant reference yaw rate $\omega_{\mathrm{ref}}$ over the optimization horizon, the integration is solved analytically, giving a linear expression $\bm{r}(t) = \bm{T}(\omega_{\mathrm{ref}},t) \bm{\xi}_i$ \ac{w.r.t.} the coefficients $\bm{\xi}_i= \begin{bmatrix} \alpha_{i,0} & \alpha_{i,1} & \alpha_{i,2} & x_{i}(\bar{t}_i) & y_{i}(\bar{t}_i) \end{bmatrix}^T$. Thus, the velocity and acceleration trajectories of spline $i$ are described by $\dot{\bm{r}}(t)=\dot{\bm{T}}(\omega_{\mathrm{ref}},t) \bm{\xi}_i$ and $\ddot{\bm{r}}(t)=\ddot{\bm{T}}(\omega_{\mathrm{ref}},t) \bm{\xi}_i$, respectively.
\subsection{Formulation of Trajectory Optimization}
To achieve robust locomotion, we deploy an online \ac{TO} which is executed in a \ac{MPC} fashion, i.e., the optimization is continuously re-evaluated providing a motion over a time horizon of $t_{\mathrm{f}}$ seconds, where $t_{\mathrm{f}}$ can be chosen as the stride duration of the locomotion gait.

The complete \ac{TO} of the wheel trajectories is formulated as a \ac{QP} problem as follows, 
\begin{equation}
\label{eq:trajectoryOptimization}
\begin{aligned}
& \underset{\bm{\xi}}{\text{min.}}
&& \frac{1}{2} \bm{\xi}^T \bm{Q}_{\mathrm{acc}} \bm{\xi} & \begin{matrix} \text{acc-} \\ \text{eleration} \end{matrix}& \\
&&& \begin{matrix*}[l] + \sum\limits_{k=1}^{N} \norm{\bm{r}(t_k) - \bm{r}_{\mathrm{pre}}(t_k+t_{\mathrm{pre}})}^{2}_{\bm{W}_{\mathrm{pre}}} \Delta t \\ \forall t \in [0,t_{\mathrm{f}}] \end{matrix*} & \begin{matrix} \text{previous} \\ \text{solution} \end{matrix}& \\
&&& \text{if leg in contact:} \\ 
&&& \hspace*{3mm} + \norm{\dot{\bm{r}}(0)-\bm{v}_{\mathrm{ref}}}^{2}_{\bm{W}_{\mathrm{ref}}} & \begin{matrix} \text{reference} \\ \text{velocity} \end{matrix}& \\
&&& \hspace*{3mm} \begin{matrix*}[l]+ \sum\limits_{k=1}^{N} \norm{r_{x}(t_k)- r_{x,\mathrm{def}}}^{2}_{w_{\mathrm{def}}} \Delta t \\ \forall t \in [\bar{t}_i,\bar{t}_i+\Delta t_i] \end{matrix*} & \begin{matrix} \text{default} \\ \text{position} \end{matrix}& \\
&&& \text{if leg in air:} \\
&&& \hspace*{3mm} + \norm{\bm{r}_{xy}(t_{\mathrm{td}})-\bm{r}_{xy,\mathrm{ref}}-\bm{r}_{xy,\mathrm{inv}}}^{2}_{\bm{W}_{\mathrm{fh}}} & \begin{matrix} \text{foothold} \\ \text{projection} \end{matrix}& \\
&&& \hspace*{3mm} + \norm{r_{z}(t_{\mathrm{sh}}) - z_{\mathrm{sh}}}^{2}_{w_{\mathrm{sh}}}, & \begin{matrix} \text{swing} \\ \text{height} \end{matrix}& \\
& \text{s.t.}
&& \bm{r}(0) = \bm{r}_{\mathrm{init}}, \dot{\bm{r}}(0) = \dot{\bm{r}}_{\mathrm{init}}, \ddot{\bm{r}}(0) = \ddot{\bm{r}}_{\mathrm{init}}, & \begin{matrix} \text{initial} \\ \text{state} \end{matrix}& \\
&&&  \begin{matrix*}[l] \begin{bmatrix*}[l] \bm{r}_{i}(\bar{t}_i+\Delta t_i) \\ \dot{\bm{r}}_{i}(\bar{t}_i+\Delta t_i) \\ \ddot{\bm{r}}_{i}(\bar{t}_i+\Delta t_i) \end{bmatrix*} = \begin{bmatrix*}[l] \bm{r}_{i+1}(\bar{t}_{i+1}) \\ \dot{\bm{r}}_{i+1}(\bar{t}_{i+1}) \\ \ddot{\bm{r}}_{i+1}(\bar{t}_{i+1}) \end{bmatrix*}, \\ \forall i \in [0,n_\mathrm{s}-1], \end{matrix*} & \begin{matrix} \text{spline} \\ \text{continuity} \end{matrix}& \\
&&& \begin{matrix*}[l] \begin{bmatrix*}[l] \abs{r_{x}(t)-r_{x,\mathrm{def}}} \\ \abs{r_{y}(t)-r_{y,\mathrm{def}}} \\ \abs{r_{z}(t)-r_{z,\mathrm{def}}} \end{bmatrix*} < \begin{bmatrix*}[l] x_{\mathrm{kin}} \\ y_{\mathrm{kin}} \\ z_{\mathrm{kin}} \end{bmatrix*}, \\ \forall t \in [0,t_{\mathrm{f}}], \end{matrix*} & \begin{matrix} \text{kinematic} \\ \text{limits} \end{matrix}& \\ 
\end{aligned}
\end{equation}
where each element is described in more detail in the following sections.
\subsection{Objectives}
\subsubsection{Acceleration minimization}
The acceleration $\ddot{\bm{r}}$ of the entire wheel trajectory is minimized to generate smooth motions and to regularize the optimization problem. The cost term for a wheel in air over the time duration $\Delta t_i$ of spline $i$ is given by
\begin{equation}
\frac{1}{2}\bm{\xi}_i^T \bigg(\underbrace{2\int_{\bar{t}_i}^{\bar{t}_i+\Delta t_i}\ddot{\bm{T}}^T(t)\bm{W}_{i,\mathrm{acc}} \ddot{\bm{T}}(t) \mathrm{d}t}_{\bm{Q}_{i,\mathrm{acc}}} \bigg)\bm{\xi}_i,
\end{equation}
where $\bm{Q}_{i,\mathrm{acc}} \in \mathbb{R}^{18 \times 18}$ is the hessian matrix, and $\bm{W}_{i,\mathrm{acc}} \in \mathbb{R}^{3 \times 3}$ is the corresponding weight matrix. Here, the linear term of (\ref{eq:quadraticProgram}) is null, i.e., $\bm{c}_{i,\mathrm{acc}}=\bm{0}_{18\times1}$. Similar, for a spline segment $i$ in contact, the hessian matrix, $\bm{Q}_{i,\mathrm{acc}} \in \mathbb{R}^{5 \times 5}$, is obtained by squaring and integrating the acceleration of the wheel trajectory over the time duration $\Delta t_i$. The time matrix $\bm{T}(\omega_{\mathrm{ref}},t)$, and hence, $\bm{Q}_{i,\mathrm{acc}}$ is dependent on the reference yaw rate as discussed in (\ref{eq:wheelSegmentSpline}).
\subsubsection{Minimize deviations from previous solution}
For a \ac{TO} with high update rates, large deviations between successive solutions can produce quivering motions. To avoid this, we add a cost term that penalizes deviations of kinematic states between consecutive solutions. We penalize the position deviations between the optimization variables from the current solution $\bm{\xi}$ and the previous solution $\bm{\xi}_{\mathrm{pre}}$ as
\begin{equation} \label{eq:minPrevSolution}
\sum\limits_{k=1}^{N} \norm{\bm{r}(t_k) - \bm{r}_{\mathrm{pre}}(t_k+t_{\mathrm{pre}})}^{2}_{\bm{W}_{\mathrm{pre}}} \Delta t, \hspace*{2mm} \forall t \in [0,t_{\mathrm{f}}],
\end{equation}
where $\bm{r}_{\mathrm{pre}}(t_k+t_{\mathrm{pre}})$ is the position vector of the wheel from the previous solution shifted by the elapsed time $t_{\mathrm{pre}}$ since computing the last solution, and $\bm{W}_{\mathrm{pre}} \in \mathbb{R}^{3 \times 3}$ is the corresponding weight matrix. This cost is penalized over the time horizon $t_{\mathrm{f}}$ with $N$ sampling points, where $t_k$ is the time at time step $k$ and $\Delta t = t_{k} - t_{k-1}$. Objectives for minimizing velocity and acceleration deviations are added in a similar formulation.
\subsubsection{Track reference velocity of wheels in contact} \label{sec:trackReferenceVel}
As shown in (\ref{eq:wheelSegmentVelocity}), the velocity along the rolling direction of the wheel trajectory is described by a quadratic polynomial which inherently satisfies the no-slip constraint. To track the reference velocity $\bm{v}_{\mathrm{ref}}$, we minimize the norm $\norm{\dot{r}_x(0)-v_{x,\mathrm{ref}}}^{2}_{w_{\mathrm{ref}}}$ which gives
\begin{equation}
\frac{1}{2} \bm{\xi}_i^T \underbrace{(2w_{\mathrm{ref}}\bm{\Gamma}^T \bm{\Gamma})}_{\bm{Q}_{i,\mathrm{ref}}} \bm{\xi}_i +  \underbrace{(-2w_{\mathrm{ref}}v_{x,\mathrm{ref}}\bm{\Gamma})}_{\bm{c}^T_{i,\mathrm{ref}}}\bm{\xi}_{i}, 
\end{equation}
where $\bm{\Gamma} = \begin{bmatrix} 1 & 0 & 0 \end{bmatrix}\dot{\bm{T}}(\omega_{\mathrm{ref}},0)$.
\subsubsection{Minimize deviations from default wheel positions}
When a wheel is in contact, differences in heading velocities of the wheels and the base can lead to configurations where the corresponding leg can get extended in the forward or backward direction. To guide the optimizer towards solutions within a desired leg configuration, we minimize the distance of the wheel from a default position $r_{x,\mathrm{def}}$ along the rolling direction $x$ as
\begin{equation} \label{eq:minDefWheelPos}
\sum\limits_{k=1}^{N} \norm{r_{x}(t_k)- r_{x,\mathrm{def}}}^{2}_{w_{\mathrm{def}}} \Delta t, \hspace*{2mm} \forall t \in [\bar{t}_i,\bar{t}_i+\Delta t_i],
\end{equation}
where $w_{\mathrm{def}}$ is the corresponding weight, and the sampling over the $i$-th contact segment's time duration $\Delta t_i$ is the same as shown in the paragraph below (\ref{eq:minPrevSolution}). 
\subsubsection{Foothold projection}
The placement of the wheel after a swing phase is crucial for hybrid locomotion (and for legged locomotion in general) because it contributes to maintaining balance and reacting to external disturbances. As shown in (\ref{eq:trajectoryOptimization}), the cost term to guide the foothold placement is given by $\norm{\bm{r}_{xy}(t_{\mathrm{td}})-\bm{r}_{xy,\mathrm{ref}}-\bm{r}_{xy,\mathrm{inv}}}^{2}_{\bm{W}_{\mathrm{fh}}}$, where $\bm{W}_{\mathrm{fh}}\in\mathbb{R}^{2\times2}$ is the weight matrix, and $t_{\mathrm{td}}=\bar{t}_i+\Delta t_i$ is the touchdown time of spline segment $i$ in air, i.e., at the end of the spline in air representing the second half of the swing phase (see Fig.~\ref{fig:wheel_trajectory}). The subscript $xy$ indicates that only footholds on the terrain plane are considered, i.e., the $z$ component is given by the height of the terrain estimation.

The position vector $\bm{r}_{xy,\mathrm{ref}}$ guides the locomotion depending on the reference velocity, which is composed of the linear velocity vector $\bm{v}_{\mathrm{ref}}$ and the angular velocity vector $\bm{\omega}_{\mathrm{ref}}$, as
\begin{equation}
\begin{bmatrix}\bm{r}_{xy,\mathrm{ref}} \\ 0 \end{bmatrix} = \begin{bmatrix}\bm{r}_{xy,\mathrm{def}} \\ 0 \end{bmatrix}  + (\bm{v}_{\mathrm{ref}} + \bm{\omega}_{\mathrm{ref}} \times \bm{r}_{BW_{xy}})\Delta t_i,
\end{equation}
where $\bm{r}_{xy,\mathrm{def}} \in \mathbb{R}^{2}$ is a specified default wheel position similar to (\ref{eq:minDefWheelPos}), and $\bm{r}_{BW_{xy}} \in \mathbb{R}^{3}$ is the position vector from the robot's \ac{COM} to the projection of the measured wheel position $W$ onto the terrain plane.

Decoupling the locomotion problem into wheel and base \ac{TO}s requires an additional heuristic to maintain balance. Balancing is achieved by adding a feedback term to the foothold obtained from reference velocities, through an inverted pendulum model~\cite{gehring2016practice,raibert1986legged} given by
\begin{equation} \label{eq:inverted_pendulum}
\bm{r}_{\mathrm{inv}} = k_{\mathrm{inv}}(\bm{v}_{BH,\mathrm{ref}} - \bm{v}_{BH})\sqrt{\frac{h}{g}},  
\end{equation}
where $\bm{v}_{BH,\mathrm{ref}} \in \mathbb{R}^3$  and $\bm{v}_{BH} \in \mathbb{R}^3$ are the reference and the measured velocity between the associated hip and base frame, respectively. Here, $h$ is the height of the hip above the ground, $g$ represents the gravitational acceleration, and $k_\mathrm{inv}$ is the gain for balancing.
\subsubsection{Swing height}
Similar to the objective in Section~\ref{sec:trackReferenceVel}, we guide the wheel \ac{TO} to match a predefined height. The objective $\norm{r_{z}(t_{\mathrm{sh}}) - z_{\mathrm{sh}}}^{2}_{w_{\mathrm{sh}}}$ given in (\ref{eq:trajectoryOptimization}) can be expanded, with a weight of $w_{\mathrm{sh}}$, to
\begin{equation}
\frac{1}{2}\bm{\xi}^T_i \underbrace{(2w_\mathrm{sh} \bm{\Gamma}^T \bm{\Gamma})}_{\bm{Q}_{i,\mathrm{sh}}}\bm{\xi}_{i} +  \underbrace{(-2w_{\mathrm{sh}} z_{\mathrm{sh}}\bm{\Gamma})}_{\bm{c}^T_{i,\mathrm{sh}}}\bm{\xi}_{i}, 
\end{equation}
with $\bm{\Gamma} = \begin{bmatrix} 0 & 0 & 1 \end{bmatrix} \bm{T}(t_{\mathrm{sh}})$, and $t_{\mathrm{sh}}=\bar{t}_i+\Delta t_i$ is the time at maximum swing height of spline segment $i$ in air, i.e., at the end of the spline in air representing the first half of the swing phase (see Fig.~\ref{fig:wheel_trajectory}).

Similarly, we set the $x$ and $y$ coordinates of the swing trajectory at maximum swing height to match the midpoint of lift-off and touch-down position. 
\subsection{Equality Constraints}
\subsubsection{Initial states}
To achieve a reactive behaviour, every optimization is initialized with the current state of the robot. As discussed in (\ref{eq:wheelSegmentSpline}), the initial position of the wheel segments in contact are set as equality constraints given by
\begin{equation}
\bm{T}(0)\bm{\xi}_{i} = \begin{bmatrix}x_{\mathrm{init}} & y_{\mathrm{init}} & 0\end{bmatrix}^T,
\end{equation}
where the initial values $x_{\mathrm{init}}$ and $y_{\mathrm{init}}$ are the measured positions of the wheel. 

If the optimization problem begins with a wheel trajectory in air, we set the initial position, velocity, and acceleration to the measured state of the wheels, i.e., $\bm{r}(0) = \bm{r}_{\mathrm{init}}$, $\dot{\bm{r}}(0) = \dot{\bm{r}}_{\mathrm{init}}$, and $\ddot{\bm{r}}(0) = \ddot{\bm{r}}_{\mathrm{init}}$.
\subsubsection{Spline continuity}
We constrain the position, velocity and acceleration at the junction of two consecutive wheel trajectory segments $i$ and $i+1$ in air as
\begin{equation} \label{eq:splineContinuity}
\begin{bmatrix}-\bm{T}_{i}(\bar{t}_i+\Delta t_i) & \bm{T}_{i+1}(\bar{t}_{i+1})\\  -\dot{\bm{T}}_{i}(\bar{t}_i+\Delta t_i) & \dot{\bm{T}}_{i+1}(\bar{t}_{i+1})\\
-\ddot{\bm{T}}_{i}(\bar{t}_i+\Delta t_i) & \ddot{\bm{T}}_{i+1}(\bar{t}_{i+1})
\end{bmatrix}
\begin{bmatrix} \bm{\xi}_{i} \\ \bm{\xi}_{i+1} \end{bmatrix} = 
\begin{bmatrix} \bm{0}_{3\times1} \\\bm{0}_{3\times1} \\\bm{0}_{3\times1} \end{bmatrix}.
\end{equation}

Junction constraints between air and contact phases are only formulated on position and velocity level. Here, the acceleration is not constrained so that the optimizer accepts abrupt changes in accelerations, allowing lift-off and touch-down events.
\subsection{Inequality Constraints}
\subsubsection{Avoid kinematic limits}
To avoid over-extensions of the legs, we keep the wheel trajectories in a kinematic feasible space which is approximated by a rectangular cuboid centered around the default positions defined in (\ref{eq:minDefWheelPos}). As introduced in (\ref{eq:trajectoryOptimization}), the kinematic limits $x_{\mathrm{kin}}$, $y_{\mathrm{kin}}$, and $z_{\mathrm{kin}}$ are enforced over the full time horizon $t_{\mathrm{f}}$ as $ \abs{r_{x}(t_k)-r_{x,\mathrm{def}}} <  x_{\mathrm{kin}}$, $\abs{r_{y}(t_k)-r_{y,\mathrm{def}}}< y_{\mathrm{kin}}$, $\abs{r_{z}(t_k)-r_{z,\mathrm{def}}}  < z_{\mathrm{kin}}$ , $\forall k \in [1,..,t_{\mathrm{f}}/\Delta t],$ with a fixed sampling time $\Delta t = t_{k} - t_{k-1}$ similar to (\ref{eq:minPrevSolution}).
\section{BASE TRAJECTORY OPTIMIZATION}
\label{sec:baseTrajectoryOptimization}
The online \ac{TO} of the base motion relies on a \ac{ZMP}~\cite{vukobratovic2004zero}-based optimization, which continuously updates reference trajectories for the free-floating base. Here, we extend the approach shown in our previous work~\cite{bjelonic2019keep}, which originates from the motion planning problem of traditional legged robots~\cite{bellicoso2017dynamic} and does not provide any optimized trajectories for the wheels/feet over a receding horizon. Moreover, the work in~\cite{bellicoso2017dynamic} only considers the optimization of the footholds. Given the wheel \ac{TO} in (\ref{eq:trajectoryOptimization}), we can generalize the idea of the \ac{ZMP} to wheeled-legged systems taking into account the trajectories of the wheels over the time horizon $t_f$.

As shown in Figure~\ref{fig:motion_planner_overview}, the motion planner of the free-floating base is described by a nonlinear optimization problem, which minimizes a nonlinear cost function $\bm{f}(\bm{\xi})$ subjected to nonlinear equality $\bm{c}(\bm{\xi})=\bm{0}$ and inequality constraints $\bm{h}(\bm{\xi})>\bm{0}$. Here, the vector of optimization variables is composed of the position of the \ac{COM} $\bm{r}_{\mathrm{COM}} \in \mathbb{R}^3$ and the yaw-pitch-roll Euler angles of the base $\bm{\theta} \in \mathbb{R}^3$.
\subsection{Parameterization of Optimization Variables and Formulation of Trajectory Optimization}
The trajectories for each \ac{DOF} of the free-floating base is represented as a sequence of quintic splines, which allows setting position, velocity and acceleration constraints. Thus, the parameterization is formulated similarly to the definition of the wheel trajectories in air given in Section~\ref{sec:wheel_segments_in_air}.

The online \ac{TO} of the base has a similar structure as the \ac{TO} described in (\ref{eq:trajectoryOptimization}). Cost terms are added to maintain smooth motions and to track the reference velocity. The equality constraints initialize the variables with the current measured state of the base and add junction constraints between consecutive splines. For balancing, we add a \ac{ZMP} inequality constraint, which is described in more detail in the next section, since this is the only part of the base optimization problem which is affected by the computed wheel trajectories in Section~\ref{sec:wheelTrajectoryOptimization}. A complete list of each objective and constraint can be obtained in~\cite{bjelonic2019keep}.
\subsection{Generalization of ZMP Inequality Constraint}
To ensure dynamic stability of the robot, the acceleration of the \ac{COM} must be chosen so that the \ac{ZMP} position $\bm{r}_{\mathrm{ZMP}} \in \mathbb{R}^3$ lies inside the support polygon\footnote{A support polygon is defined by the convex hull of the expected wheels' contact trajectories.}. This nonlinear inequality constraint is given by
\begin{equation} \label{eq:zmp_constraint}
\begin{bmatrix}
p(t_k) & q(t_k) & 0
\end{bmatrix}
 \bm{r}_{\mathrm{ZMP}}(t_k)+r(t_k) \geq 0, \hspace*{2mm}\forall t_k\in[0,t_{\mathrm{f}}]
\end{equation}
where $\bm{r}_{\mathrm{ZMP}} = \bm{n} \times \bm{m}_{\mathrm{gi}}/ (\bm{n}^T \bm{f}_{\mathrm{gi}})$~\cite{sardain2004forces} and $\bm{n} \in \mathbb{R}^3$ is the the terrain normal. The \emph{gravito-inertial} wrench~\cite{caron2017zmp} is given by $\bm{f}_{\mathrm{gi}} = m \cdot (\bm{g} - \ddot{\bm{r}}_{\mathrm{COM}}) \in \mathbb{R}^3$ and $\bm{m}_{\mathrm{gi}} = m \cdot \bm{r}_{\mathrm{COM}} \times (\bm{g}-\ddot{\bm{r}}_{\mathrm{COM}}) - \dot{\bm{l}}_{\mathrm{COM}} \in \mathbb{R}^3$, where $m$ is the mass of the robot, $\bm{l}_{\mathrm{COM}} \in \mathbb{R}^3$ is the angular momentum of the \ac{COM}, and $\bm{g} \in \mathbb{R}^3$ is the gravity vector. In contrast to \cite{bellicoso2017dynamic,bjelonic2019keep}, the line coefficients $\bm{d}(t)=[p(t) \ q(t) \ r(t)]^T$ that describe an edge of a support polygon depend on the time $t$, since the contact points of wheeled-legged robots continue to move even when a leg is in contact, unlike conventional legged robots. The \ac{ZMP} inequality constraint is sampled over the time horizon $t_{\mathrm{f}}$ with a fixed sampling time $\Delta t = t_k - t_{k-1}$.
\section{EXPERIMENTAL RESULTS AND DISCUSSION}
\label{sec:experiments}
To validate the performance of our hybrid locomotion framework, this section reports on experiments and real-world applications conducted with ANYmal equipped with non-steerable, torque-controlled wheels (see Fig.~\ref{fig:anymal_on_wheels}). A video\footnote{\hbox{Available at \href{https://youtu.be/ukY0vyM-yfY}{https://youtu.be/ukY0vyM-yfY}}} showing the results accompanies this paper.

\subsection{Implementation}
The wheel \ac{TO}, base \ac{TO}, tracking controller, and state estimator are running on a single PC (Intel i7-7500U, 2.7 GHz, dual-core 64-bit). All computation regarding the autonomy, i.e., perception, mapping, localization, path planning, path following, and object detection, is carried out by three different PCs. The robot is entirely self-contained in terms of computation and perception. As can be obtained in Fig.~\ref{fig:motion_planner_overview}, we run each wheel \ac{TO}, the base \ac{TO}, and the \ac{WBC} in concurrent threads where each optimization reads the last available solutions from its predecessor. Moreover, all optimization problems are run online due to fast solver times.

A hierarchical \ac{WBC} tracks the computed trajectories in Section~\ref{sec:wheelTrajectoryOptimization} and Section~\ref{sec:baseTrajectoryOptimization} by generating torque commands for each actuator and accounting for the full rigid body dynamics including its physical constraints, e.g., the non-holonomic rolling constraint, friction cone, and torque limits~\cite{bjelonic2019keep}. The \ac{WBC} runs together with state estimation~\cite{bloesch2018two} in a \unit[400]{Hz} loop. Similar to~\cite{bloesch2013stateconsistent}, we fuse the \ac{IMU} reading and the kinematic measurements from each actuator to acquire the robot's state. Moreover, the frame $W$ in Fig.~\ref{fig:wheel_trajectory} requires an estimate of the terrain normal. In this work, the robot is locally modeling the terrain as a three-dimensional plane, which is estimated by fitting a plane through the most recent contact locations~\cite{bjelonic2019keep}. The contact state of each leg is determined through an estimation of the contact force, which takes into account the measurements of the motor drives and the full-rigid body dynamics.

We model and compute the kinematics and dynamics of the robot based on the open-source \ac{RBDL}~\cite{rbdl}, which uses the algorithms described in~\cite{featherstone2014rigid}. The nonlinear optimization problem in Section~\ref{sec:baseTrajectoryOptimization} is solved with a custom \ac{SQP} algorithm, which solves the problem by iterating through a sequence of \ac{QP} problems. Each \ac{QP} problem including the optimization problem in Section~\ref{sec:wheelTrajectoryOptimization} is solved using QuadProg++~\cite{quadProg}, which internally implements the Goldfarb-Idnani active-set method~\cite{goldfarb1983numerically}. To maintain a positive definite Hessian $\bm{Q}$ in (\ref{eq:quadraticProgram}) and to ensure the convexity of the resulting \ac{QP} problem, a regularizer $\rho$ is added to its diagonal elements, e.g., $\rho = 10^{-8}$ as in~\cite{bellicoso2017dynamic}. The tuning of the cost function in (\ref{eq:trajectoryOptimization}) remains a manual task where a single value describes the diagonal elements of the weighting matrices, and one parameter set is provided for all motions shown next.
\subsection{Solver Time of Different Contact Scheduler and Gait Switching}
As shown in Table~\ref{table:solver_time}, the wheel and base optimizations are solved in the order of milliseconds, and a great variety of gaits from driving, i.e., all legs in contact, up to gaits with full-flight phases are possible. Besides, the accompanying video shows manual gait switches between driving and hybrid walking-driving gaits, which can be useful for future works regarding automatic gait switches to reduce the \ac{COT} further.
\begin{table}[b]
    \caption{Time horizon $t_{\mathrm{f}}$ and optimization times including model setup for different gaits. The reported solver times for wheel TO are for one wheel, and the hybrid running trot is a gait with full-flight phases.}
    \label{table:solver_time}
    \begin{center}
    \begin{tabular}{cccc}
    \toprule
    Gait &$t_{\mathrm{f}}$ / (s) &Wheel \ac{TO} / (ms) &Base \ac{TO} / (ms) \\\midrule
    Driving &1.7 &0.14 & 6.93 \\
    Hybrid walk &2.0 &0.81 & 14.83 \\
    Hybrid pace &0.95 &0.42 & 1.88 \\
    Hybrid trot &0.85 &0.47 & 2.4 \\ 
    Hybrid running trot &0.64 &0.58 & 5.77
    \\\bottomrule
    \end{tabular}
    \end{center}
\end{table}
\subsection{Rough Terrain Negotiation}
\begin{figure}[t]
    \centering
    \includegraphics[width=\columnwidth]{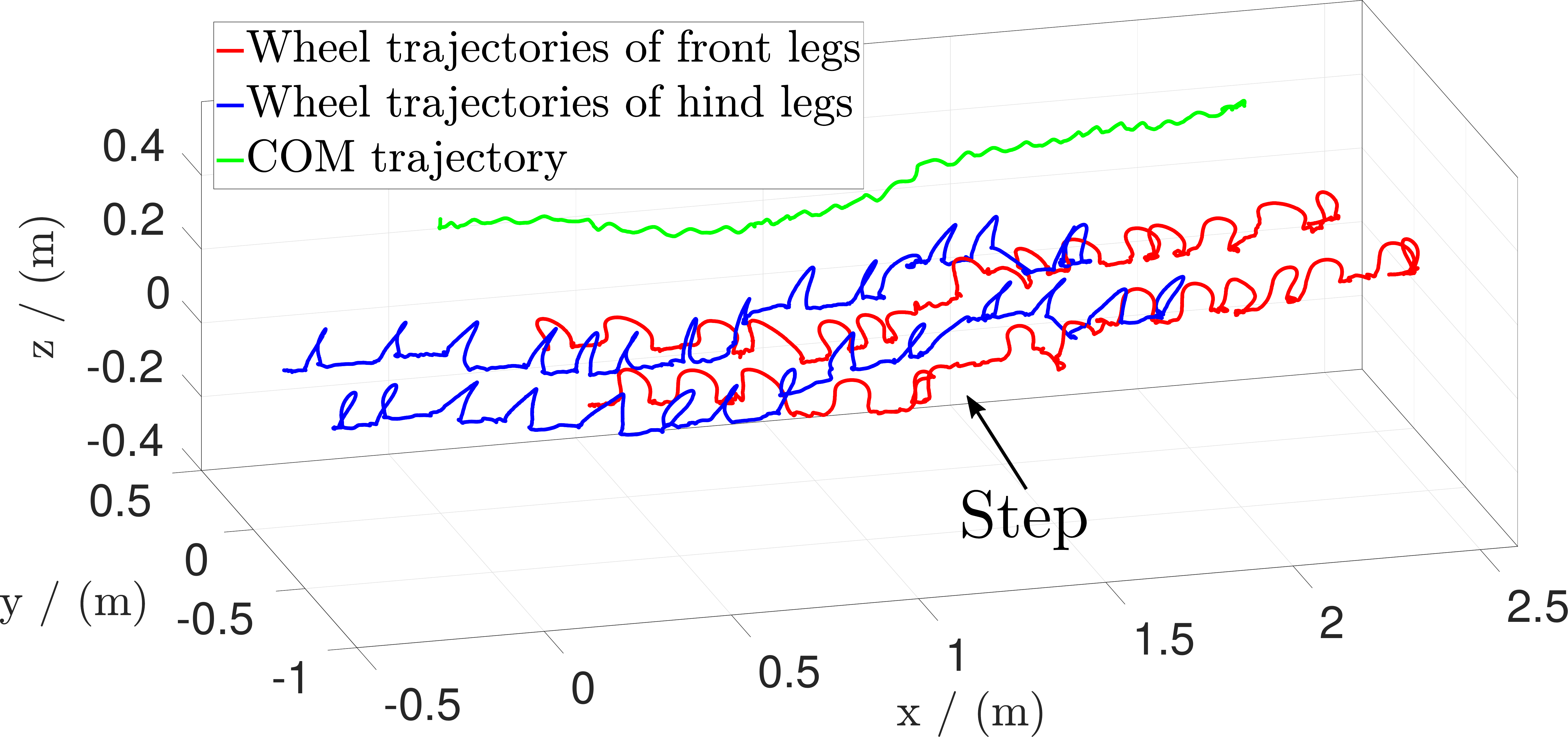}
    \caption{Measured \ac{COM} and wheel trajectories of ANYmal over a step while hybrid trotting, as depicted in the upper images of Fig.~\ref{fig:anymal_on_wheels}. The three-dimensional plot shows the wheel trajectories of the front legs (red line), the wheel trajectories of the hind legs (blue line), and the \ac{COM} trajectory (green line) \ac{w.r.t.} the inertial frame, which is initialized at the beginning of the run.}
    \label{fig:step_3d_plot}
\end{figure}

The robot is capable of blind locomotion in a great variety of unstructured terrains, e.g., inclines, steps, gravel, mud, and puddles. Fig.~\ref{fig:anymal_on_wheels} and the accompanying video shows the performance of the robot in these kinds of environments. As depicted in Fig.~\ref{fig:step_3d_plot}, the robot can overcome blindly steps up to \unit[20]{\%} of its leg length. The obstacle verifies the advantage of our hybrid locomotion framework. In contrast to the related work and our previous work~\cite{bjelonic2019keep}, the robot traverses obstacles without stopping and switching to a pure walking motion. To our best knowledge, this is the first time a robot has demonstrated this level of obstacle negotiation at high speeds, with multiple gaits. Moreover, the locomotion becomes more robust since the framework accounts for possible motions on the ground. The accompanying video shows an instance where the wheel collides with the edge of a step. Our framework is capable of adapting to these scenarios by merely driving over the obstacle.
\subsection{High Speed and Cost of Transport}
On flat terrain, the robot achieves a mechanical \ac{COT}~\cite{bjelonic2018skating} of 0.2 while hybrid trotting at the speed of \unit[2]{m/s} and the mechanical power consumption is \unit[156]{W}. The \ac{COT} is by a factor of two higher than a pure driving gait at the same speed. A comparison to traditional walking and skating with passive wheels~\cite{bjelonic2018skating} shows that the \ac{COT} is lower by \unit[42]{\%} \ac{w.r.t.} the traditional trotting gait and by \unit[9]{\%} \ac{w.r.t.} skating motions.
\subsection{DARPA Subterranean Challenge: Tunnel Circuit}
The first \ac{DARPA} Subterranean Challenge, the Tunnel Circuit, was held close to Pittsburgh in the NIOSH mine. The main objective was to search, detect, and provide autonomously spatially referenced locations of artifacts inside the underground mine. The wheeled version of ANYmal participated in two runs as part of the CERBERUS team~\cite{cerberus} alongside flying and other mobile platforms. Moreover, the wheeled quadrupedal robot was deployed next to the traditional version of ANYmal without wheels.

\begin{figure}[t]
    \centering
    \includegraphics[width=\columnwidth]{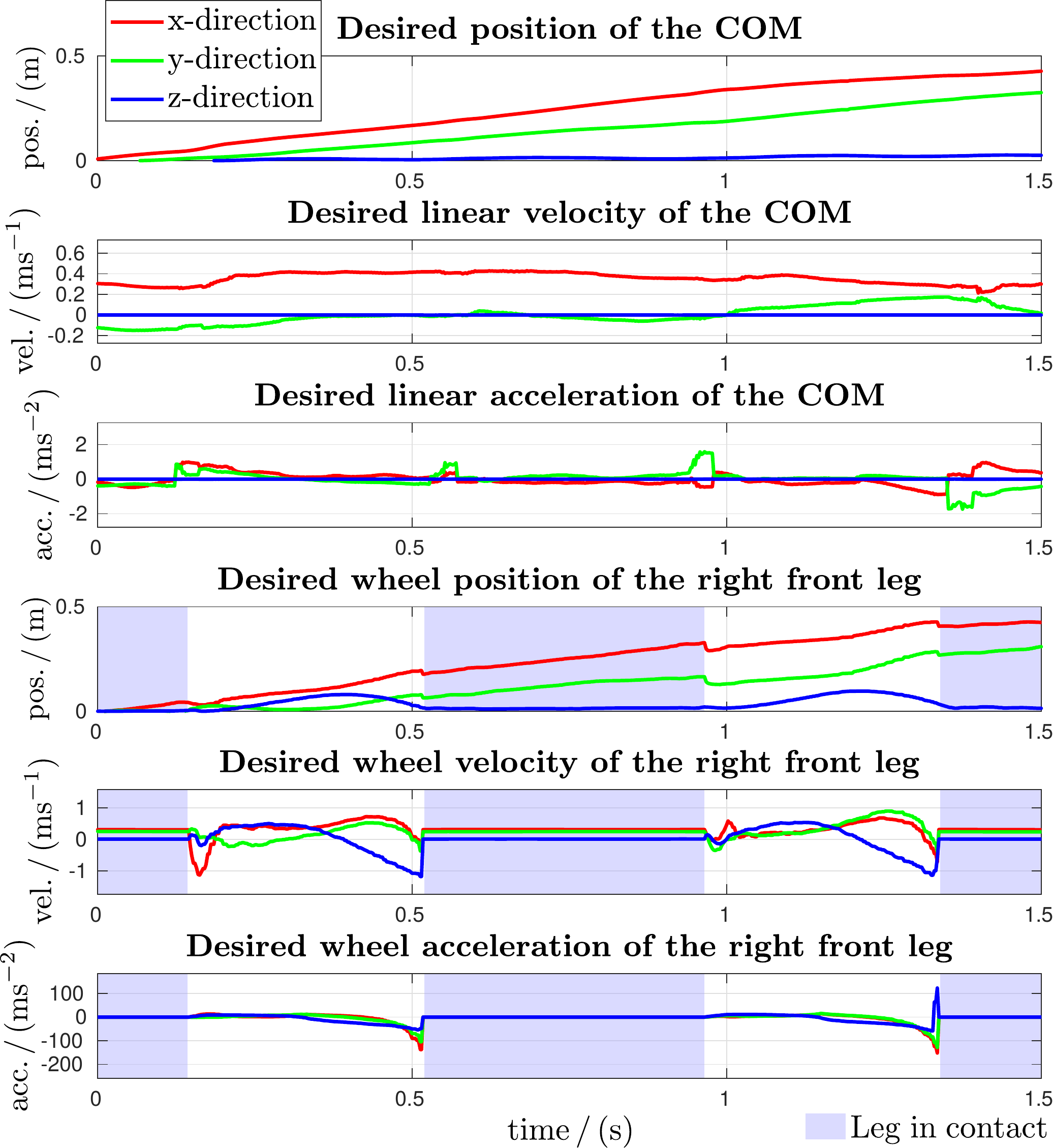}
    \caption{Desired \ac{COM} and wheel trajectories of ANYmal at the DARPA Subterranean Challenge. The robot, ANYmal, is autonomously locomoting with a hybrid driving-trotting gait during the second scoring run. The environment is a wet, inclined, muddy, and rough underground mine, as depicted in the lower images of Fig.~\ref{fig:anymal_on_wheels}. Despite the challenging terrain, the robot manages to explore fully autonomously the mine for more than \unit[100]{m}. The plots show the desired motions for approximately two stride durations. Due to the fast update rates of the \ac{TO} problems and reinitialization of the optimization problem with the measured state, the executed trajectories are almost identical to the desired motion shown here.}
    \label{fig:darpa_2d_plot}
\end{figure}

As depicted in the lower images of Fig.~\ref{fig:anymal_on_wheels}, the terrain consisted of hilly, bumpy, and muddy terrain and in some parts of the mine, the robot needed to cross puddles. Throughout both runs, the robot locomoted the terrain with a hybrid trot. In the first run, the wheeled version of ANYmal traversed \unit[70]{m} without significant issues, and the robot successfully reported the correct location of one artifact. In the end, however, one of the wheels started slipping on the muddy terrain before the fall. As can be seen in the accompanying video, the robot managed to balance after the first slip because of the foothold adaptation of the inverted pendulum model in (\ref{eq:inverted_pendulum}). The mechanical design was improved after the first run by adding a chain around the wheels to increase the friction coefficient while traversing the mud (see the right middle image of Fig.~\ref{fig:anymal_on_wheels}). Fig.~\ref{fig:darpa_2d_plot} shows the desired trajectories of the \ac{COM} and wheels for a few meters of the second run. Here, it can be seen that the robot executes a hybrid trotting gait since, during ground contact, the wheel moves along its rolling direction. Despite the challenging environment, the hybrid locomotion framework enabled the robot to travel for more than \unit[100]{m}.

Due to the time limitation of the challenge, the speed of mobile platforms becomes an essential factor. Most of the wheeled platforms shown from the other competing teams were faster than our traditional legged robot by a factor of two or more. The upcoming Urban Circuit of the Subterranean Challenge includes stairs and other challenging obstacles. Therefore, we believe, only a wheeled-legged robot is capable of combining speed and versatility. At the Tunnel Circuit, the wheeled version of ANYmal traversed with an average speed of \unit[0.5]{m/s}, which was more than double the average speed of the traditional legged system. Our chosen speed was limited by the update frequency of our mapping approach or otherwise could have traversed the entire terrain with much higher speeds without any loss in agility. On the whole, the performance validation for real-world applications is satisfying, and a direct comparison with the traditional ANYmal reveals the advantages of wheeled-legged robots.
\section{CONCLUSIONS}
\label{sec:conclusions}
This work presents an online \ac{TO} generating hybrid walking-driving motions on a wheeled quadrupedal robot. The optimization problem is broken down into wheel and base trajectory generation. The two independent \ac{TO}s are synchronized to generate feasible motions by time sampling the prior generated wheel trajectories, which form the support polygons of the \ac{ZMP} inequality constraint of the base \ac{TO}. The presented algorithm makes the locomotion planning for high dimensional wheeled-legged robots more tractable, enables us to solve the problem in real-time on-board in a \ac{MPC} fashion, and increases the robustness in the robot's locomotion against unforeseen disturbances.

To the best of our knowledge, this is the first time that a hybrid walking-driving robot is deployed for real-world missions at one of the biggest robotics competition. In future work, we plan to incorporate the optimization of the gait timings to enable automatic switching between pure driving and hybrid walking-driving. As shown in our work, an automated way of choosing when to lift a leg can increase the speed and robustness of the locomotion.
%







\balance
\bibliographystyle{IEEEtran}
\bibliography{IEEEabrv,submissionbibfile}

\end{document}